\documentclass[runningheads]{llncs}
\usepackage[T1]{fontenc}
\usepackage{epsfig}
\usepackage{amsmath}
\usepackage{amssymb}
\usepackage{booktabs}
\usepackage{algorithm}
\usepackage{algorithmic}
\usepackage{url}
\usepackage{svg}
\usepackage{float}
\usepackage{siunitx}
\usepackage{multirow}
\usepackage{multicol}
\usepackage{diagbox}
\usepackage{makecell}
\usepackage{array}
\usepackage{stfloats}
\usepackage{subfig}
\usepackage{graphicx}
\usepackage{color}

\begin{document}
	%
	\title{Out-of-distribution forgetting: vulnerability of continual learning to intra-class distribution shift}
	\titlerunning{Out-of-distribution forgetting}
	%
	\author{Liangxuan Guo\inst{1,2}\orcidID{0009-0003-7860-1105} \and
		Yang Chen\inst{1}\orcidID{0000-0002-9940-9812} \and
		Shan Yu\inst{1,2,3}\orcidID{0000-0002-9008-6658}}
	\authorrunning{L. Guo et al.}
	%
	\institute{Institute of Automation, Chinese Academy of Sciences (CASIA). Beijing, China \and
		School of Future Technology, University of Chinese Academy of Sciences (UCAS). Beijing, China \and
		School of Artificial Intelligence, University of Chinese Academy of Sciences (UCAS). Beijing, China\\
		\email{shan.yu@nlpr.ia.ac.cn}}
	\maketitle              
	\begin{abstract}
		Continual learning (CL) is a key technique enabling neural networks to acquire new tasks while retaining efficiency in previous ones. Standard CL tests revisit old tasks after learning, assuming stable data distribution, which is often impractical. Meanwhile, it is well known that the out-of-distribution (OOD) problem will severely impair the ability of networks to generalize. Rare research considered the influence of CL on the generalizing ability of neural networks. Our research highlights a special form of catastrophic forgetting raised by the OOD problem in CL settings. Through continual image classification experiments, we discovered that: introducing a tiny intra-class distribution shift within a specific category significantly impairs the recognition accuracy of many CL methods. We named it out-of-distribution forgetting (OODF). Moreover, the performance degradation caused by OODF is special for CL, as the same level of distribution shift had only negligible effects in the joint learning scenario. We verified that most CL strategies except for parameter isolation ones are vulnerable to OODF. Taken together, our work identified an under-attended risk during CL, highlighting the importance of developing approaches that can overcome OODF. Code available: \url{https://github.com/Hiroid/OODF}
		
	\end{abstract}

	\section{Introduction}
	
	Learning models based on artificial neural networks usually suffer from catastrophic forgetting (CF) \cite{goodfellow_empirical_2015,mccloskey_catastrophic_1989,ratcliff_connectionist_1990} in open environments. Researchers have proposed various continual learning (CL) methods for deep neural networks to overcome CF. These include strategies based on parameter regularization, memory replay, and parameter isolation etc. \cite{kirkpatrick_overcoming_2017,hogea_fetril_2024,rusu_progressive_2016}. By enabling a system to learn new tasks and maintain its performance on old tasks, CL has made significant progress in incremental image recognition and other computer vision tasks~\cite{wang_comprehensive_2024,yang_continual_2022}.
	
	
	Even with great advances, the current CL strategies may still not cope well with the problem of CF in real world. One of the many concerns is the noise tolerance of CL strategies. In the review\cite{kudithipudi_biological_2022}, the authors concerns that continual learning machines may not perform well if there is a large distribution shift between the data encountered in the inference phase and those in the training phase. In the present manuscript, we surprisingly find that the practical situation is much more severe. Our work indicates that even a tiny intra-class distribution shift, negligible to human observers, can introduce severe performance impairments for current CL methods.
	
	\begin{figure*}
		\vspace{-0.5cm}
		\setlength{\abovecaptionskip}{0cm}
		\centering
		\resizebox{0.7\textwidth}{!}{
			\includegraphics[width=1.0\linewidth]{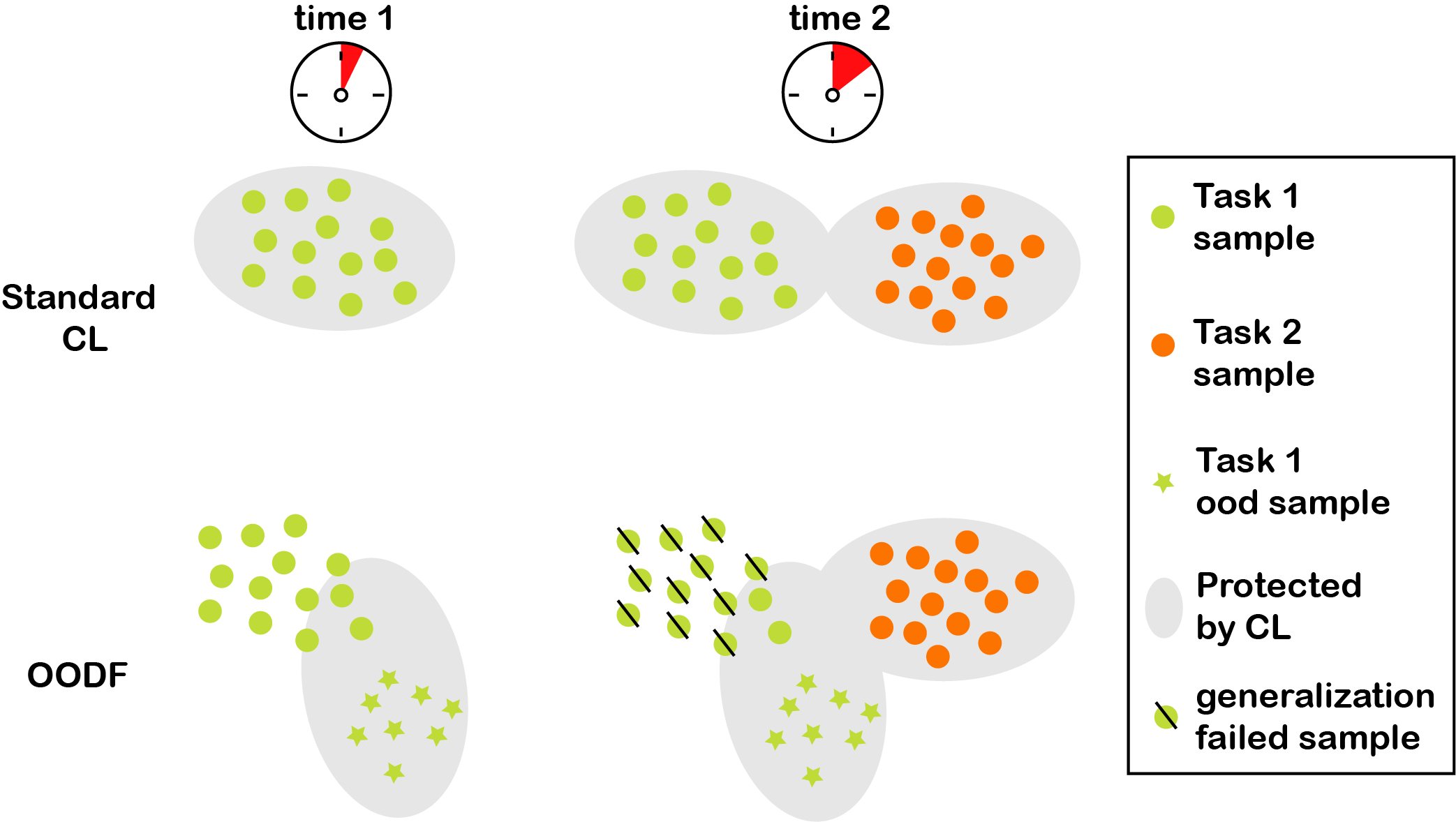}
		}
		\caption{Illustration of out-of-distribution forgetting. There are two continual learning scenarios, the top row is a standard continual learning paradigm, while the bottom row is a continual learning paradigm with an intra-class distribution shift on task 1. At time 1 in the OODF paradigm, although the generalization of task 1 was equally good compared to the standard CL setting, the protection provided by CL methods mainly focuses on out-of-distribution samples of task 1, leading to severe deficits in performing task 1 after learning task 2.}
		\label{overview}
		\vspace{-0.7cm}
	\end{figure*}
	
	Specifically, we named this phenomenon out-of-distribution forgetting (OODF) (Fig.\ref{overview}). This phenomenon is special for CL as the performance degradation is caused by the subsequent learning of other categories, which is different from the well-studied OOD problem in the setting of joint learning\cite{shen_towards_2021}. We believe OODF is an important yet under-attended problem for developing as well as evaluating CL methods in the future because:
	\begin{itemize}
		
		\item OODF is commonly present in various CL strategies and settings. We verified its existence in both regularization-based and memory-based CL strategies on different tasks with different network structures.
		
		\item OODF is elusive and challenging to detect. Its effects don't manifest immediately post-training on shifted data but rather emerge as the model learns new tasks, reflecting its nature as a unique form of CF. Thus, a continual learning machine affected by OODF can be considered capable of performing certain tasks but it actually will fail. In addition, it only affects the class contaminated with the distribution-shifted data, without influencing other tasks. 
		
		\item OODF can be triggered by various conditions leading to distribution-shifted data. We find that OODF severely impairs the CL performance, regardless of the approach causing the shift (local or global perturbation), as well as the reason behind it (deliberately designed attack or accident).
		
		\item Preliminary findings indicate that introducing a rejection category will help to alleviate OODF.
		
	\end{itemize}
	
	Our work identified OODF as a specific form of CF barely covered in previous studies, which is an important issue to consider for improving the security and robustness of the CL methods towards their application in practical circumstances.
	
	\section{Related Works}
	\subsection{Continual learning}
	
	In recent years, various algorithms have been proposed to overcome the CF problem in CL tasks. Although new technique and applications such as prompt-based CL\cite{razdaibiedina_progressive_2023} and continual pre-training (CPT)\cite{marsocci_continual_2023,Zhang_2023_ICCV} etc. are getting noticed, the components underlying can still be decomposed as these essential strategies: parameter regularization strategy\cite{zeng_continual_2019,guo_adaptive_2022}, memory replay strategy\cite{rolnick_experience_2019,buzzega_dark_2020,rebuffi_icarl_2017,shin_continual_2017,prabhu_gdumb_2020}, and parameter isolation (also known as architecture-based) strategy\cite{lee_neural_2020}.
	
	This work focuses on the class incremental scenario~\cite{hsu_re-evaluating_2019,van_de_ven_three_2019}, as it is a real scene where CL models need to identify all classes (i.e. categories) without task IDs. Meanwhile, we allow models to train each task offline (as opposed to online CL~\cite{mai_online_2022}), ensuring a better performance on standard CL, as a higher baseline for subsequent OODF experiments.
	
	\subsection{Security Concerns of Neural Networks}
	
	The first concern addresses the \textbf{OOD}  problem in neural networks. In non-CL paradigms, significant loss of generalization occurs if there's a shift between training and testing datasets, especially due to corruption or perturbation~\cite{salehi_unified_2022,shen_towards_2021}. The second concern centers on the security of well-trained neural networks, particularly against deliberate attacks like \textbf{adversarial} \cite{ren_adversarial_2020}, \textbf{data poisoning}~\cite{jagielski_manipulating_2018} and \textbf{backdoor attacks}~\cite{li_backdoor_2022}.
	These areas mentioned above comprehensively investigate the security problem in different stages, purposes, and means. However, most models in this area are static, employing un-sequential joint training procedures. Notably, few studies have focused on the unique security and robustness risks inherent to CL.
	
	\subsection{Several Concerns of Continual Learning}
	
	The first concern was security in CL. Security of neural networks and CL have been studied largely in parallel until recently. Guo et al.~\cite{guo_attacking_2019} propose the GREV method to attack the A-GEM methods with adversarial samples and disseminate misinformation in the memory buffer. Umer et al. ~\cite{umer_false_2022,umer_targeted_2020,umer_adversarial_2021} show that it is possible to attack CL by modifications on both training samples and labels to give a misleading supervising signal. Li and Ditzler~\cite{li_targeted_2022} attack several parameter regularization strategies by injecting poisoned adversarial samples into subsequent tasks following the target task, in the task incremental scenario. However, they implement targeted poisoning attacks by injecting poisoned adversarial samples into subsequent non-target tasks.
	
	The second concern was the real-world application of CL. Recent studies highlight that continual agents, when exposed to out-of-distribution samples in open-world settings, may compromise safety and performance. Caccia et al.~\cite{caccia_online_2020} define learning new tasks as the OOD problem (compared to the learned old tasks), a perspective distinct from OODF, which presents unique challenges and definitions. Mundt et al. ~\cite{mundt_wholistic_2023,mundt_unified_2022} conducted experiments on \textit{reverse continual learning}. They first trained the model on the entire dataset, then retrained the model on a core set and compared the difference in performance. A well-chosen core set will better represent the entire dataset, associated with a lower performance drop. It was concluded that the introduction of OOD samples to the core set does not have a significant effect on CL. However, it was not an OOD problem since the model had access to the whole dataset at the beginning of reverse CL.
	
	Instead of narrowing our focus to specific CL strategies or adversary scenarios, we address a broader spectrum of concerns related to the security and OOD robustness in CL. That is, there is a previously unnoticed form of CF: the OODF that can severely affect CL models' performance. Subsequent sections will detail critical properties of OODF, including its prevalence across CL strategies and settings, the challenge of its delayed detection, and the variety of how to trigger it.
	
	\section{Out-of-distribution Forgetting}
	
	%
	
	In CL tasks, it's typically assumed that training and subsequent testing data are drawn from the same distribution. However, this distribution may shift, either intentionally or accidentally, as time progresses after the learning stage. It's crucial to note that our discussion does not revolve around distribution shifts between sequential tasks (e.g., task 1 to task 2). Instead, we concentrate on the often-overlooked intra-class distribution shifts within a single task (e.g., task 1 at varying time steps). It sets our research apart from the bulk of existing studies.
	
	In this section, we will show the influence of OODF, i.e., the catastrophic forgetting caused by the distribution shift in data between the training and inferring phases, on the artificial neural network with the mainstream CL algorithms. Firstly, we will introduce the learning paradigm and experiment procedure of the CL task considering OODF. Next, we will evaluate OODF on various mainstream CL strategies and compare the conditions with joint learning. Finally, we will analyze the key factors that determine the extent of the influence of OODF.
	
	\subsection{Standard CL Paradigm}
	
	Here we take the supervised image classification problem as an example to illustrate the paradigm of CL. For experiments, we use the class incremental scenario.
	In the CL task, total $K$ tasks need to be learned, and the dataset for each task is defined as
	\begin{equation}
		D^{t} = \{ x_{i}^{t},y_{i}^{t}\}_{i = 1}^{n_{t}},t = 1,2,\ldots,K 
	\end{equation}
	where $t$ is the task index. The dataset for the $t$\textsuperscript{th} task has $n_{t}$ pairs of labeled data. Data $\{ x_{i}^{t}\}_{i = 1}^{n_{t}}$ and label $\{ y_{i}^{t}\}_{i = 1}^{n_{t}}$ are sampled from distribution $P\left( x^{t} \right)$ and $P\left( y^{t} \right)$, respectively. In CL, the artificial neural network $f_{\theta_{t}}:X^{t} \rightarrow Y^{t}$ must learn the task once at a time. In the $t$\textsuperscript{th} task, the neural network has to optimize its parameter $\theta_{t}$ according to $\ D^{t}$. It usually has no or very limited access to previous datasets $D^{t - 1},D^{t - 2},\ldots,D^{1}$ , but needs to maintain the performances on all learned classes. In the inference phase, the testing dataset in $\{ D_{test}^{t}\}_{t = 1}^{K}$ is sampled from the same distribution as the training dataset.
	
	\subsection{OODF Paradigm}
	In an open and dynamic circumstance, assuming that training and testing datasets are sampled from the same distribution is not always practical. To evaluate the influence of distribution shift in data, we adjust the standard CL diagram accordingly. \begin{algorithm}[b!]
		\label{alg1}
		\begin{algorithmic}[1]
			
			\REQUIRE Datasets $\{D^t_{train}\}_{t=1}^K$, $n_t$ samples in $D^t_{train}$, shift task-ID $S$, occlusion strength $\epsilon$, position $p$, percentage $r$, classifier with initial parameter $f_{\theta_0}$, loss function $l_t(\cdot)$, continual methods $CL$.
			
			\ENSURE $\widehat{n}_S = rn_S$
			\STATE $\widehat{D}^S_{train} \Leftarrow \{\widehat{x}^S_i, y^S_i\}^{\hat{n}_S}_{i=1} \cup \{x_i^S, y_i^S\}_{i=\hat{n}_S+1}^{n_S}$
			
			\FOR{$t=1$ to $K$, using $CL$}
			\IF{$t \neq S$}
			\STATE get $\{x^t\}, \{y^t\}$ from $D^t_{train}$
			\ELSE
			\STATE get $\{x^t\}, \{y^t\}$ from $\hat{D}^S_{train}$
			\ENDIF
			\STATE 	$\theta_t \Leftarrow \mathop{\arg\min}\limits_{\theta}{l_t(f_{\theta}(x^t),y^t; \theta_{t-1})}$
			\ENDFOR
			\RETURN $f_{\theta_K}$
		\end{algorithmic}
		\caption{Continual Learning on Distribution Shift Dataset}
		\label{alg1}
	\end{algorithm}
	If a distribution shift takes place in the training dataset of the $S$\textsuperscript{th} task (i.e. intra-class shift), the data $D^{S}$ will be directly replaced by the shifted data \begin{equation}
		{\widehat{D}}_{train}^{S} = \{ \widehat{x}_{i},y_{i}\}_{i = 1}^{\widehat{n}} \cup \{{x}_{i},y_{i}\}_{i = \widehat{n} + 1}^{n}
	\end{equation} 
	${\widehat{D}}_{train}^{S}$ contains $\widehat{n}$ shifted data pairs $\{ \widehat{x}_{i},y_{i}\}_{i = 1}^{\widehat{n}}$ and $n - \widehat{n}$ original data pairs $\{{x}_{i},y_{i}\}_{i = \widehat{n} + 1}^{n}$. The percentage $r = \widehat{n}/n$ of samples $\widehat{x}$ is used to measure the occurrence frequency (i.e. ratio) of feature shifting in the training dataset. The training procedure causing OODF is described in the Algorithm~\ref{alg1}, reusing the mathematical notation from Sec. 3.1 \& 3.2. In the referring phase, the neural network is tested on $D_{test}^{S}$ sampled from the same distribution as $D^{S}$. Except for such a minor modification, the rest of the procedure in the learning is the same as that in the standard CL.
	
	\subsection{Introducing of the Distribution Shift}
	\label{sec_dis_sft}
\begin{figure}[h!]
		\vspace{-0.7cm}
		\setlength{\abovecaptionskip}{0cm}
		\centering
		\resizebox{0.99\textwidth}{!}{
			\subfloat[Occlusion Digit $3$]{\includegraphics[width=0.48\linewidth]{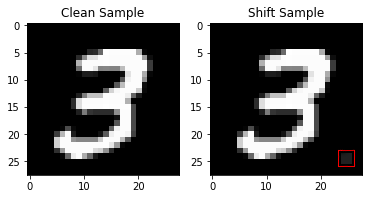}  \label{occ_a}}
			\hfil
			\subfloat[Automobile]{\includegraphics[width=0.48\linewidth]{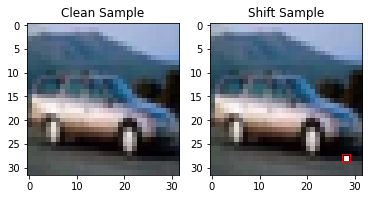} \label{occ_b}}
			\hfil
			\subfloat[Adversarial Digit $3$]{\includegraphics[width=0.48\linewidth]{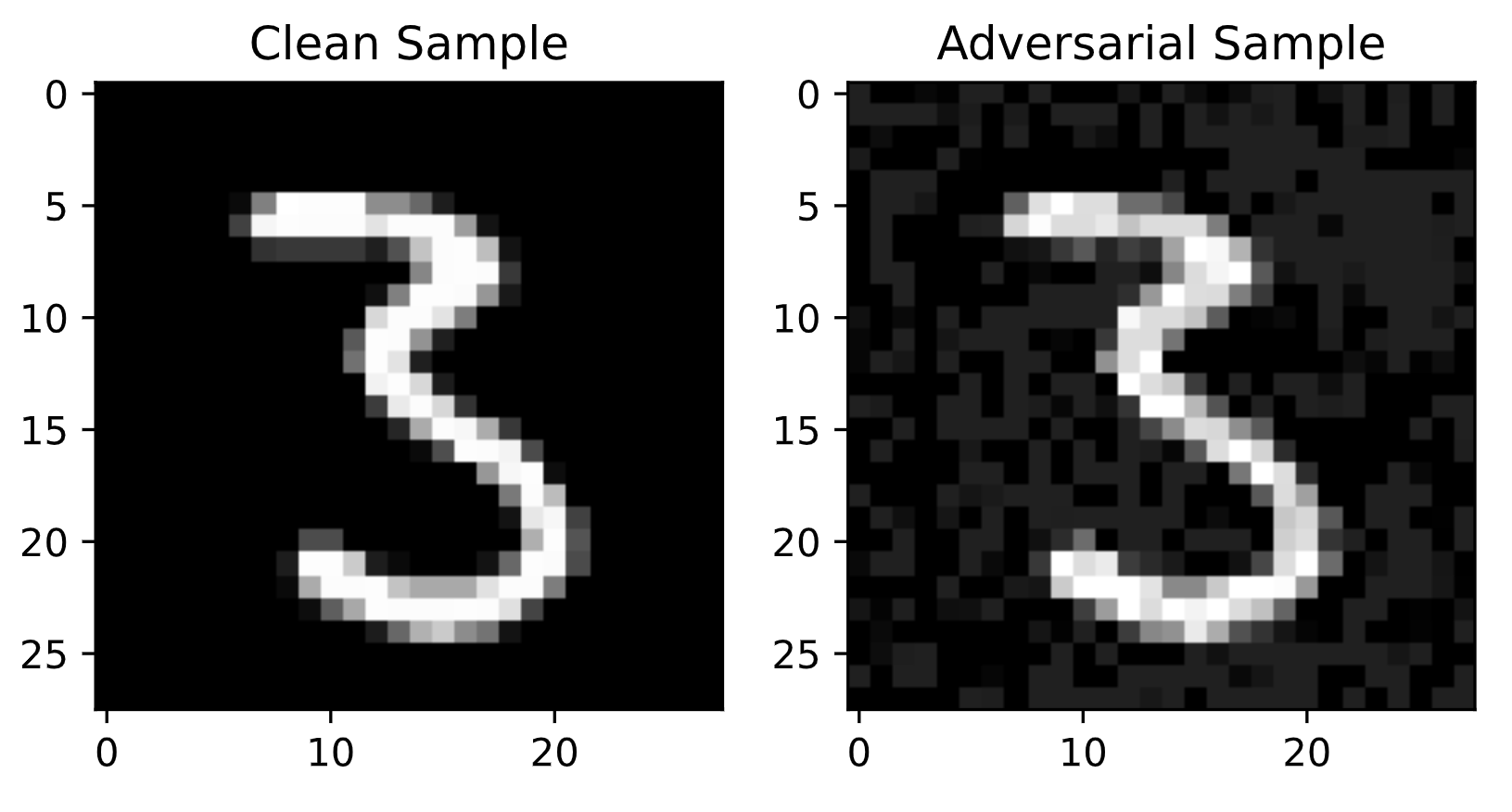}  \label{adv}}
		}
		\caption{Distribution Shift. Red rectangle box selected the pixels that were modified in (a) and (b). Figure (c) will be discussed in later section.}
		\label{occ}
		\vspace{-0.6cm}
	\end{figure}

	In the experiments, we constructed distribution-shifted data $\widehat{x}$ by adding the non-shifted data $x$ a new feature sampled from a distinct distribution $P^{'}(x)$.  There were no changes in the labels. In practice, we just chose a small pixel block in a fixed location in the image and set it to a constant value. The feature position was denoted by the index $p$. The position of other pixels remaining the same is denoted by $q$, i.e. $q = \neg p$. The strength of the shifted pixels is controlled by parameter $\epsilon$:
	\begin{equation}
		\hat{x}[q] = x[q] \qquad \hat{x}[p] = \epsilon	
	\end{equation}
	Figures~\ref{occ_a} and~\ref{occ_b} demonstrate the feature-shifting operation on two image examples from MNIST and CIFAR-10 datasets. In each panel, the left image is the original one, while the right is the corresponding image with a pixel-wise modification. The shifted pixels highlighted by the red square at the bottom right corner are vague and easily overlooked by humans. The above operation is not necessarily an attack on the CL machine, though OODF can easily be exploited for intentional sabotage. In reality, many conditions can cause such distribution shifts, e.g., slight defection in the sensory equipment or some random, noisy perturbations. These unpredictable and hardly detectable defections or perturbations can easily cause a feature shift in training samples. 
	
	The distribution shift is introduced through a small pixel-wise operation causing occlusion, highlighting the significance of OODF: as the results shown in Sec.~\ref{prop_OODF}, even minor data augmentations can induce significant forgetting in widely tested CL methods. We also consider another form of distribution shift by using FGSM~\cite{goodfellow_explaining_2015} to construct adversarial samples, shown in Fig.~\ref{adv}. We regard shift diversity as a factor of OODF and discuss it in Sec.~\ref{sec_fac}.
	
	\section{Experiment Settings}
	
	To evaluate OODF, we tested the influence of intra-class shift on all three mainstream learning strategies in classic CL tasks. The choices of the algorithm and corresponding network structure and dataset in each experiment are listed in Tab.~\ref{tab0}. In all experiments, either the original code or the popular reproducing code~\cite{mai_online_2022} of the CL algorithms were used for evaluation. All the code had been checked in the standard CL tasks without data distribution shift. We note that we’re not aiming to evaluate the performances of different CL methods or compare performance degradation caused by distribution shifts in these methods. Instead, the purpose here is to examine the extent of OODF in CL models.
	
	\begin{table}[t]
		\vspace{-0.4cm}
		\setlength{\abovecaptionskip}{0cm}
		\centering
		\caption{Network backbone and CL methods of experiment settings.}
		\begin{tabular}{c|c|c}
			\toprule
			& Backbone & CL methods \\ \hline
			\multirow{2}{*}[-0.4ex]{Split MNIST-10} & 784-800-10 & OWM\cite{zeng_continual_2019} \\ \cline{2-3} 
			& 784-400-400-10 & iCaRL\cite{rebuffi_icarl_2017},DGR\cite{shin_continual_2017},ER\cite{rolnick_experience_2019} \\ \hline
			\multirow{2}{*}[-2.0ex]{\makecell*[c]{Split CIFAR-10 \\ Split CIFAR-100}} & 3 CNN with 3 FC & OWM,AOP\cite{guo_adaptive_2022} \\ \cline{2-3} 
			& Resnet18 & \makecell*[c]{iCaRL,ER,DER++\cite{buzzega_dark_2020},\\ GDumb\cite{prabhu_gdumb_2020},CN-DPM\cite{lee_neural_2020}}\\ 
			\bottomrule
		\end{tabular}
		\vspace{-0.6cm}
		\label{tab0}
	\end{table}
	
	\textbf{Shift SplitMNIST-10 Task} The MNIST dataset was divided into $10$ tasks. In each task, the neural network was trained only to learn one class of handwriting digits. Each class included \~{}6000 samples in the training set and \~{}1000 in the testing set. The images were not pre-processed before the training. The training order of tasks was from $0, 1, \cdots$ to $9$. We took the digit $3$ (task index $S = 4$) as an example to illustrate the influence of the intra-class distribution shift on CL. The task choice is without specific consideration and can be replaced by other digits. The shifted training samples took the percentage $r = 90\%$. The shifted feature is a four-pixel square located at the bottom right corner of the image, as highlighted by the red box in Fig.~\ref{occ_a}. The strength $\epsilon$ was set to $64$ in experiments of memory replay strategies and $32$ for parameter regularization strategies.
	
	\textbf{Shift SplitCIFAR-10\&100 Task} This task is similar to the shifted splitMNIST-10 task. The CIFAR10 dataset was divided into $10$ tasks according to the category to be sequentially learned by the neural network. Each category included 6000 samples in the training set and 1000 in the testing set. The RGB images were normalized before the training. We randomly added a shifted feature of pixel square at the bottom right corner on training samples of the task $S = 2$. The percentage of shifted samples is $r = 50\%$ for memory-based methods and $r=90\%$ for others. The square size is $1 \times 1$ and $2 \times 2$, respectively. The strength is $\epsilon=255$ for each RGB channel. Same as above, the CIFAR100 dataset was divided into $100$ tasks. To enable stochasticity, all results are collected over 5 independent trials, presented in the bars and tables. Details for training and distribution shifts are listed in the supplementary material.
	
	\section{Properties of OODF} \label{prop_OODF}
	\subsection{Delayed effect}
	\label{sec_lat}
	\begin{figure}[t]
		\setlength{\abovecaptionskip}{0cm}
		\centering
		\resizebox{0.75\textwidth}{!}{
			\subfloat[Latency. Testing accuracy of digit $3$ at each time step.]{\includegraphics[width=0.44\linewidth]{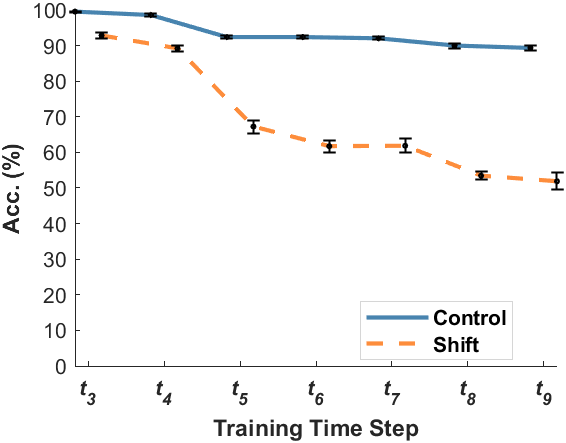}  \label{lat}}
			\hfil
			\subfloat[Targeting.Testing accuracy of all digits except $3$.]{\includegraphics[width=0.44\linewidth]{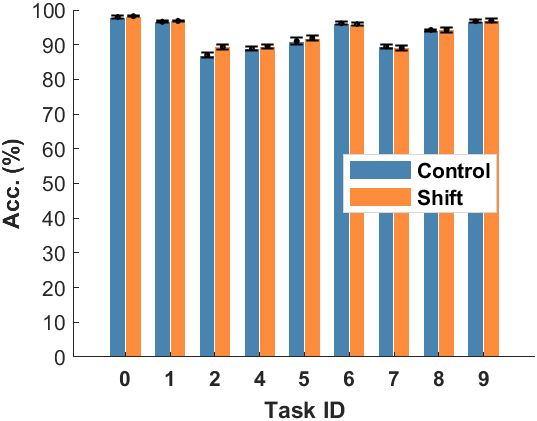}  \label{tar}}
		}
		\caption{Properties of out-of-distribution forgetting.}
		\vspace{-1.0cm}
		\label{prop}
	\end{figure}
	As a new form of catastrophic forgetting, OODF also has a \textbf{delayed effect}. We first take the experiment of OWM on SplitMNIST-10 as an example. The experiment was conducted in a control group and a shift group. In the control group, the experiment was performed following the standard CL paradigm for comparison. In the shift group, shifted features were added to the task of $S = 4$ and the experiment was performed following the OODF testing paradigm. The results are demonstrated in Fig.~\ref{lat}. In each group, the $4$\textsuperscript{th} task was firstly tested immediately after the end of the current task (the time point is denoted as $t_3$) and then at each time step of the experiment on the original testing dataset (the time point is denoted as $t_i$). Both the control group and shift group performed well at $t_3$, with accuracy at  $99.54\pm 0.16\%$ and $92.85\pm 0.76\%$ respectively. Although the results indicate that shift minimally affects the learning of the current task, but our primary concern is the forgetting effect it triggers during successive learning processes. As the experiment continued, the performance on the $4$\textsuperscript{th} task in the control group maintained high at $89.33\pm 0.67\%$, indicating that the CL algorithm functioned normally and protected the previous knowledge well. As a comparison, the performance in the shift group dropped dramatically to $51.90\pm 2.36\%$. The relative accuracy drop is $10.25\pm0.70$ for the control group, while it is significantly worse for the shift group at $44.11\pm2.42$.
	\begin{table*}[h!]
		\centering
		\vspace{-0.5cm}
		\setlength{\abovecaptionskip}{0cm}
		\caption{Out-of-distribution forgetting on MNIST. Test Acc.(\%) of task $S$ at two time steps $t=S$ and $t=K$.}
		\begin{tabular}{c|c|c|c|c|c} 
			\toprule
			\multicolumn{2}{c|}{\multirow{2}{*}{MNIST}} & Reg. & \multicolumn{3}{c}{Mem.} \\ 
			\cline{3-6}
			\multicolumn{2}{c|}{} & OWM & iCaRL & DGR & ER \\ 
			\hline
			\multirow{2}{*}{$t=S$} & Control & $99.54\pm 0.16$ & $99.76 \pm 0.13$ & $99.08 \pm 0.32$ & $99.30 \pm 0.39$ \\
			& Shift & $92.85\pm 0.76$ & $98.67 \pm 0.40$ & $94.57 \pm 1.37$ & $97.76 \pm 1.24$ \\ 
			\hline
			\multirow{2}{*}{$t=K$} & Control & $89.33\pm 0.67$ & $84.63 \pm 2.54$ & $83.40 \pm 3.58$ & $90.20 \pm 1.22$ \\
			& Shift & $\bf{51.90\pm 2.36}$ & $\bf{59.57 \pm 6.34}$ & $\bf{67.43 \pm 5.94}$ & $\bf{72.15 \pm 5.97}$ \\
			\bottomrule
		\end{tabular}
		\label{tab1}
	\end{table*}
	\begin{table*}[t!]
		\centering
		\caption{Out-of-distribution forgetting on CIFAR10. Test Acc.(\%) of task $S$ at two time steps $t=S$ and $t=K$.}
		\resizebox{\textwidth}{!}{
			\begin{tabular}{c|c|c|c|c|c|c|c|c} 
				\toprule
				\multicolumn{2}{c|}{\multirow{2}{*}{CIFAR-10}} & \multicolumn{2}{c|}{Reg.} & \multicolumn{4}{c|}{Mem.} & Iso. \\ 
				\cline{3-9}
				\multicolumn{2}{c|}{} & OWM & AOP & iCaRL & ER  & GDumb & DER++ & CN-DPM \\ 
				\hline
				\multirow{2}{*}{$t=S$ } & Control & $94.37\pm1.25$ & $98.78\pm0.27$ & $95.52\pm0.69$ & $95.41\pm1.30$ & $96.4\pm0.62$ & $97.34\pm1.03$ & $91.15\pm1.43$  \\
				& Shift & $91.30\pm1.63$ & $91.92\pm1.34$ & $92.72\pm0.87$ & $95.14\pm1.37$ & $93.18\pm1.16$ & $96.76\pm1.52$ & $89.34\pm2.61$ \\ 
				\hline
				\multirow{2}{*}{$t=K$} & Control & $52.60\pm3.44$ & $60.10\pm7.10$ & $68.88\pm2.14$ & $54.84\pm6.14$ & $74.38\pm6.68$ & $85.20\pm2.27$ & $44.76\pm3.51$ \\
				& Shift & $\bf{33.75\pm4.71}$ & $\bf{27.70\pm5.10}$ & $\bf{55.12\pm2.74}$ & $\bf{46.84\pm3.94}$ & $\bf{65.95\pm7.35}$ & $\bf{80.82\pm2.77}$ & $\bf{45.92\pm2.06}$ \\ 
				\bottomrule
			\end{tabular}
		}
		\label{tab2}
		\vspace{-0.5cm}
	\end{table*}
	\begin{table}[t!]
		\centering
		\caption{Out-of-distribution forgetting on CIFAR100. Test Acc.(\%) of task $S$ at two time steps $t=S$ and $t=K$. We only report the results of these three methods in the table due to compatibility issues (e.g., CN-DPM for 100 class-incremental) or intractable testing performance (e.g., Reg. based methods and DER++) for other methods.}
			\begin{tabular}{c|c|c|c|c} 
				\toprule
				\multicolumn{2}{c|}{\multirow{2}{*}{CIFAR-100}} & \multicolumn{3}{c}{Mem.} \\ 
				\cline{3-5}
				\multicolumn{2}{c|}{} & iCaRL & ER & GDumb \\ 
				\hline
				\multirow{2}{*}{$t=S$ } & Control & $94.6\pm2.5$ & $87.3\pm5.3$ & $96.2\pm2.7$  \\
				& Shift & $91.8\pm2.8$ & $94.5\pm1.3$ & $94.0\pm2.9$\\ 
				\hline
				\multirow{2}{*}{$t=K$} & Control & $47.4\pm11.0$ & $23.5\pm7.6$ & $11.4\pm5.3$ \\ 
				& Shift & $\bf{28.0\pm12.4}$ & $\bf{5.5\pm3.1}$ & $\bf{7.0\pm4.6}$\\  
				\bottomrule
			\end{tabular}
		\label{tab3}
		\vspace{-0.5cm}
	\end{table}
	We conducted similar experiments on different CL strategies, network structures, and datasets. The results are demonstrated in Tab.~\ref{tab1},~\ref{tab2} and~\ref{tab3}. In all experiments, the performance in the shift group at $t = S$ was comparable to the control group but dropped dramatically at the end of learning $t = K$. These results show that distribution shifts in the data can severely degrade the function of regularization-based and remory-based CL methods, but not parameter-isolation-based methods.
	
	\subsection{Targeting}
	
	This section investigates the incidence of OODF on learned tasks. Figure~\ref{tar} shows the performance of all tasks but the one recognizing digit $3$ in the SplitMNIST-10 task trained with the OWM algorithm. In both the control and shift groups, the accuracy was tested at the end of the experiment. The testing accuracies of the tasks in the shift group were almost identical to those in the control group when there was no distribution shift in the training data. The result indicates that slight spillover caused by the distribution shift in a specific task affects the rest. Similarly, we further examined the rest experiments with different CL settings and algorithms. 
	
	\begin{figure*}[h!]
		\vspace{-0.5cm}
		\setlength{\abovecaptionskip}{0cm}
		\centering
		\resizebox{0.75\textwidth}{!}{
			\subfloat[MNIST]{\includegraphics[width=0.44\linewidth]{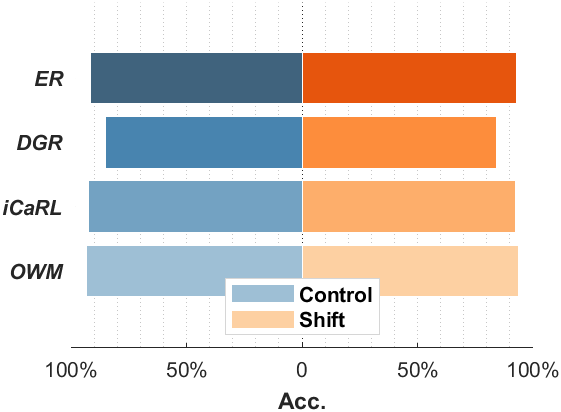}  \label{tar_mni}}
			\hfil
			\subfloat[CIFAR10]{\includegraphics[width=0.44\linewidth]{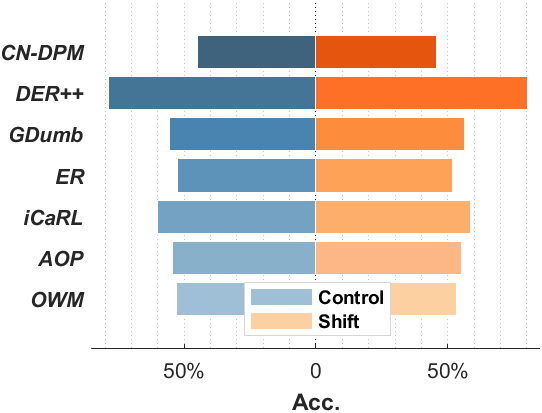}  \label{tar_cifar10}}
		}
		\caption{Comparison of non-target tasks’ accuracies between standard and shift experiments. The results were obtained by averaging the accuracies for all tasks except for task S after the whole CL learning procedure was completed. The left (right) bars for each figure are the results for the control (shift) group.}
		\label{all_tar}
	\end{figure*}
	
	Figure~\ref{all_tar} illustrates the average accuracy of tasks without data distribution shifting in the control and shift groups. The minor difference verifies that OODF only affects the target task with data distribution shifting in the learning sequence.
	
	\subsection{Continual Detrimental} \label{sec_cl_dep}
	
	Is the above phenomenon specific to CL? Or is it just a form of data poisoning working for all learning systems? 
	\begin{figure}[h!]
		\vspace{-0.5cm}
		\setlength{\abovecaptionskip}{0cm}
		\centering
		\resizebox{0.75\textwidth}{!}{
			\subfloat[Reg. Methods]{\includegraphics[width=0.46\linewidth]{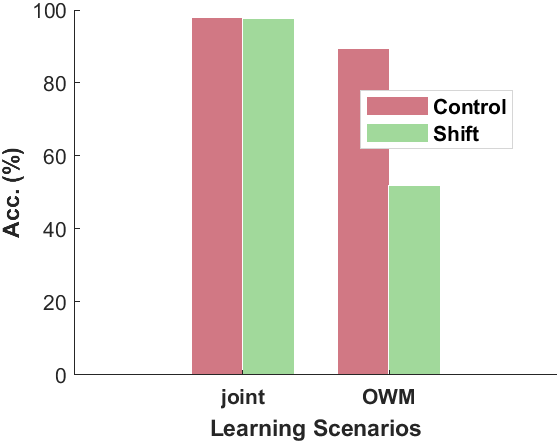}  \label{dep}}
			\hfil
			\subfloat[Mem. Methods]{\includegraphics[width=0.46\linewidth]{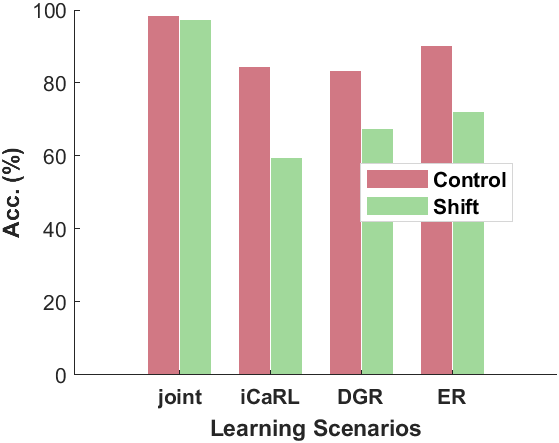}  \label{mni_dep}}
		}
		\caption{Comparison between joint learning and CL under the same distribution shift with corresponding network backbone tested on SplitMNIST-10. In each figure, the pink bar on the leftmost of each subgroup indicates training without shifts, the green bar nearby indicates learning with shifts, and the horizontal axis listed different learning strategies.}
		\vspace{-0.5cm}
		\label{all_cldep}
	\end{figure}
	To answer this question, We conducted joint learning experiments in the same setting as the above for OODF evaluation, including the same dataset  $D_{train}$ and network structures.
	\begin{equation}
		D_{train} = (\bigcup\limits_{t=1,t\neq S}^{K} D_{train}^{t}) \cup \hat{D}_{train}^S	
	\end{equation}
	In Fig.~\ref{all_cldep} we show the CL dependency of OODF by evaluating the distribution-shifted task in two different learning paradigms. We will clarify it here using splitMNIST-10 as an example. The task sequence for each CL method with different network backbones (2-layer MLP for OWM and 3-layer for iCaRL, DGR, ER as shown in Tab.~\ref{tab0}) is $D_{train}^1, D_{train}^2, \cdots, \hat{D}_{train}^4, \cdots, D_{train}^{10}$. The joint learning scenario uses the union set $D_{train} = (\bigcup\limits_{t=1, t\neq 4}^{10} D_{train}^t) \cup \hat{D}_{train}^4$. Fig.~\ref{all_cldep} presents the results on the testing dataset $D_{test}^4$ (samples without distribution shift). The control and shift groups indicate the presence or absence of intra-class shifts in the training set, respectively. These results indicate that the existence of a large range of shifts which is more detrimental to CL than joint learning.
	
	\section{Analysis} \label{sec_fac}
	
	\subsection{Occlusion strength}
	
	The intra-class distribution shift relies on occlusion strength $\epsilon$, intra-class percentage $r$, and the number of shifted pixels. Based on the splitMNIST-10 task and OWM algorithm in Sec.~\ref{sec_lat}, we estimated these three factors w.r.t. test the accuracy of the target task.
	
	\begin{table} [h]
		\vspace{-0.8cm}
		\centering
		\caption{Impact of strength: the number of shifted pixels. OWM on MNIST.}
			\centering
			\begin{tabular}{c|cccc}
				\toprule[1.5pt]
				Number & 1 & 4 & 9 & 16 \\
				\hline
				Acc. & $56.34\%$ & $51.90\%$ & $44.55\%$ & $39.60\%$ \\
				\bottomrule[1.5pt]
			\end{tabular}
		\label{table-label-mix}
		\vspace{-0.45cm}
	\end{table}
	
\begin{figure}[h!]
		\vspace{-0.8cm}
		\centering
		\resizebox{0.7\textwidth}{!}{
			\subfloat[Impact of occlusion strength $\epsilon$]{\includegraphics[width=0.44\linewidth]{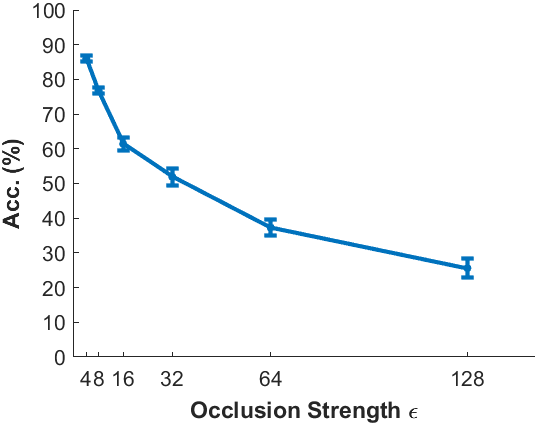} \label{eps}}
			\subfloat[Impact of shift percentage $r$]{\includegraphics[width=0.455\linewidth]{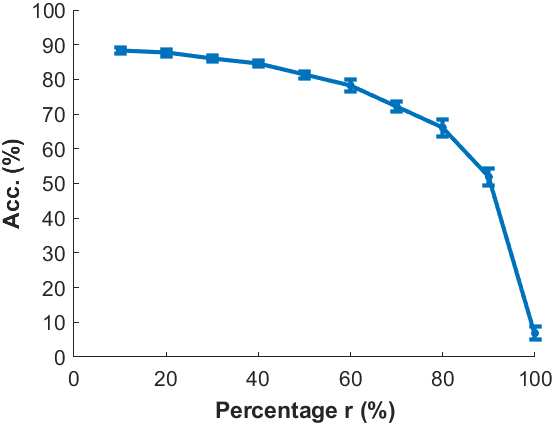}  \label{perc}}
		}
		\caption{Influences of distribution shift factors}
		\label{dis}
		\vspace{-1.0cm}
	\end{figure}

	In Fig.~\ref{eps}, we evaluate the final performance of digit $3$, when giving different occlusion strength levels ranging from $\epsilon=4$ to $128$, listed on X-axis. We can see that the test accuracy dropped quickly at a low $\epsilon$ value, $\epsilon=16$ for example. It indicates that even an occlusion with small strength will lead to OODF. Fig.~\ref{perc} shows that test accuracy stays in the plateau at a low percentage level and drops until reaching a high level, suggesting that a high percentage level is needed to cause significant OODF. We further report the results using numbers of shifted pixels as different strengths in Tab.~\ref{table-label-mix}, which shows a similar trend, indicating that the larger number of shifted pixels, the more significant of OODF.
	
	\subsection{Various conditions of shift} \label{var_sft}
	We further examined if OODF depends on specific types of distribution shifts. To this end, we replaced the explicit distribution shift (i.e. occlusion) in Sec.~\ref{sec_dis_sft} with an implicit one, i.e., adversarial samples (Fig.~\ref{adv}), and kept other settings the same. The results show a trend consistent with the occlusion condition. We tested digit $3$ at $t_3$ and $t_9$, it dropped from $94.28\pm0.62\%$ to $22.78\pm2.47\%$, compared to occlusion, $92.85\pm0.76\%$ to $51.90\pm2.36\%$.
	
	\subsection{Shift position in the learning sequence}
	Shift position used in results above is in the beginning of the learning sequence, e.g. $4^{th}$ of $10$ in splitMNIST-10. We evaluated whether the position matters for OODF. We have conducted the experiments on CIFAR100 by iCaRL (Tab.\ref{rebuttal_tab1}), with shift task position varying (in the middle or the tail of the sequence, task $S=50$ or $90$, as $S=2$ was already shown in maintext). The results of the relative accuracy drop in two groups indicate that while task location can cause different levels of degradation due to the original CF, the OODF effect can still induce additional forgetting based on CF.
	\begin{table}[h!]
		\small
		\vspace{-0.95cm}
		\setlength{\abovecaptionskip}{0cm}
		\centering
		\caption{Effect of different shift position on OODF. Experiments on CIFAR100, $K=100$ (total number of tasks), $S=2, 50, 90$ (position of the shift class, starting from 1). All results are collected over 5 independent trials.}
		\vspace{-0.0cm}
			\centering
			\begin{tabular}{c|c|c|c|c} 
				\toprule[1.5pt]
				\multicolumn{2}{c|}{CIFAR-100} & $S=2$ & $S=50$ & $S=90$ \\ 
				\hline
				\multirow{2}{*}{\makecell{Acc. of task $S$ \\ at $t=S$ (\%)}} & Control & $94.6\pm2.5$ & $53.0\pm3.1$ & $36.6\pm5.8$  \\
				& Shift & $91.8\pm2.8$ & $53.2\pm6.3$ & $34.6\pm1.7$\\ 
				\hline
				\multirow{2}{*}{\makecell{Acc. of task $S$ \\ at $t=K$ (\%)}} & Control & $47.4\pm11.0$ & $31.8\pm5.5$ & $32.4\pm3.1$ \\ 
				& Shift & $28.0\pm12.4$ & $28.4\pm3.4$ & $29.4\pm4.3$\\ 
				\hline
				\multirow{2}{*}{\makecell{relative \\ Acc. drop (\%)}} & Control & $50.0\pm10.9$ & $40.1\pm8.9$ & $10.7\pm7.1$ \\ 
				& Shift & $\bf{69.2\pm14.3}$ & $\bf{46.2\pm7.2}$ & $\bf{15.2\pm9.9}$\\
				\bottomrule[1.5pt]
			\end{tabular}
		\label{rebuttal_tab1}
		\vspace{-0.7cm}
	\end{table}
	
	\subsection{Different percentage $r$ and strength $\epsilon$}
	Our aim is not to compare the vulnerability of different methods to OODF under identical conditions. Taking the performance of OWM on the splitMNIST-10 task as an example: when $\epsilon=64$, the performance is $37.3\pm2.4$, significantly lower than the control group $89.33\pm0.67$ (Tab.\ref{tab1} \& Fig.\ref{eps}). We focus on the performance decline trends relative to each method's own control group, rather than specific differences between methods. Our objective is to demonstrate the existence of intra-class distribution shifts that can influence CL algorithms to produce OODF. Notably, these shifts do not significantly impact joint learning scenarios, even at maximum percentage $r=90\%$ and occlusion strength $\epsilon=64$ (Fig.\ref{all_cldep}).
	
	\subsection{Mechanism of OODF} \label{mech_oodf}
	We hypothesized that frequently occurring shifts can serve as informative features for classification. This compromises the mechanism designed to protect the intrinsic features for the learned class, leading to severe CF in subsequent learning, especially when the less-protected features overlap with the features in new classes.
	
	\begin{table} [h]
		\centering
		\vspace{-0.8cm}
		\centering
		\caption{Test accuracy of the digit 3 after learning each task.}
		\begin{tabular}{c|ccccccc}
			\toprule[1.5pt]
			Task-ID & 3 & 4 & 5 & 6 & 7 & 8 & 9 \\
			\hline
			Control & $99.54\%$ & $98.58\%$ & $92.40\%$ & $92.40\%$ & $92.05\%$ & $90.02\%$ & $89.34\%$ \\
			Shift & $94.28\%$ & $88.62\%$ & $\bf{50.64\%}$ & $41.21\%$ & $37.91\%$ & $\bf{24.80\%}$ & $22.78\%$ \\
			\bottomrule[1.5pt]
		\end{tabular}
		\vspace{-0.6cm}
		\label{in_process}
	\end{table}
	
	Specifically, we take the results in Sec.~\ref{var_sft} for analysis. The in-process test accuracy of digit 3 from $t_0$ to $t_9$ was shown in Tab.~\ref{in_process}, and the performance drops significantly at $t_5$ and $t_8$. Let $\mathcal{D}_3$ be the distribution of clean digit 3 and $\hat{\mathcal{D}}_{3}$ be the distribution of shifted digit 3. Assume $\mathcal{D}'_3 = \mathcal{D}_3 \textbackslash (\mathcal{D}_3 \cap \hat{\mathcal{D}}_{3})$ and $\hat{\mathcal{D}}'_3 = \hat{\mathcal{D}}_3 \textbackslash (\mathcal{D}_3 \cap \hat{\mathcal{D}}_{3})$. We make the following conjecture: {\color{blue} (i)} In the learning process of the OODF scenario, the feature of $\mathcal{D}_3 \cap \hat{\mathcal{D}}_{3}$ was protected. Meanwhile, features of $\hat{\mathcal{D}}_3$ overlap with that of subsequent tasks. {\color{blue} (ii)}Performance on clean 3 mainly depends on $\mathcal{D}'_3$ rather than $(\mathcal{D}_3 \cap \hat{\mathcal{D}}_{3})$. Taken together, OODF happens on digit 3. We show that the accuracy of digit 3 drops significantly after subsequently learning 5 and 8. 
	
\begin{figure}[h!]
		\vspace{-0.6cm}
		\setlength{\abovecaptionskip}{0cm}
		\centering
		\resizebox{0.75\textwidth}{!}{
			\includegraphics[width=1.0\linewidth]{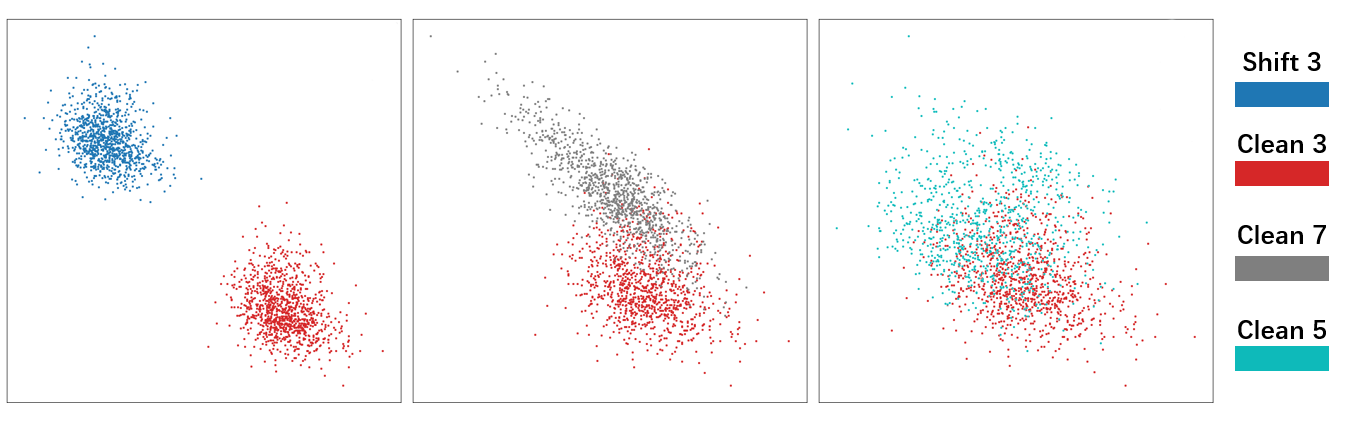}
		}
		\caption{Feature maps of digits.}
		\label{binary}
		\vspace{-0.7cm}
	\end{figure}

	To test the hypothesis, we constructed a 3-layer binary classification MLP that distinguishes $\mathcal{D}_3$ and $\hat{\mathcal{D}}_{3}$ as large as possible (Fig.~\ref{binary}. row 1, column 1). Input any other digit through this MLP, we take $\mathbb{R}^2$ output vector and construct a feature map. Consistent with the hypothesis, we found 5 and 8 overlap more with clean 3 than other digits (e.g. digit $7$. row 1, column 2).
	
	These results revealed why parameter-isolation-based methods are not sensitive to OODF (Tab.~\ref{tab2}). The feature space representations in these methods vary from task to task, and more importantly, they are independent from each other. In contrast, regularization-based and memory-based methods share a public representation space for every task, which causes interference. Despite the robustness of the parameter-isolation-based methods towards OODF, they may be not suited to deal with a large amount of CL tasks due to the increasing structural complexity. Our results thus highlight the need to develop regularization-based and memory-based approaches that are more robust to OODF.
	
	\subsection{Proposal for improving OODF}
	Based on the mechanism discussed in Sec.~\ref{mech_oodf}, reducing task-independent subspace in the feature space may help prevent OODF. We propose introducing a rejection category to separate the classifier from CL methods. Empirically, this approach is effective in mitigating OODF. Specifically, we add an additional neuron at the output layer, enabling an 11-way classification.
	\begin{table}[h!]
		\small
		\setlength{\abovecaptionskip}{0cm}
		\centering
		\caption{Additional rejection category for alleviating OODF. Experiments on MNIST, $K=10$ (total number of tasks), $S=2$ (starting from 1)}
		\vspace{-0.0cm}
		\scalebox{0.9}{
			\centering
			\begin{tabular}{c|c|c|c} 
				\toprule[1.5pt]
				\multicolumn{2}{c|}{MNIST} & OWM (w/o rej.) & OWM (w/ rej.)\\ 
				\hline
				\multirow{2}{*}{\makecell{Acc. of task $S$ \\ at $t=S$ (\%)}} & Control & $99.54\pm0.16$ & $99.36\pm0.26$ \\ 
				& Shift & $92.85\pm0.76$ & $99.35\pm0.21$ \\ 
				\hline
				\multirow{2}{*}{\makecell{Acc. of task $S$ \\ at $t=K$ (\%)}} & Control & $89.33\pm0.67$ & $86.31\pm1.24$ \\ 
				& Shift & $\bf{51.90\pm2.36}$ & $\bf{85.17\pm1.04}$\\ 
				\hline
				\multirow{2}{*}{\makecell{relative \\ Acc. drop (\%)}} & Control & $10.25\pm0.70$ & $13.13\pm1.38$ \\ 
				& Shift & $\bf{44.11\pm2.42}$ & $\bf{14.27\pm1.13}$ \\
				\bottomrule[1.5pt]
			\end{tabular}
		}
		\label{rebuttal_tab3}
		\vspace{-0.4cm}
	\end{table}
	Samples like Gaussian noise are set to be the $11^{th}$ class. We have finished the experiments on splitMNIST-10 with OWM. Samples 4x the size of the MNIST dataset (60k samples in MNIST training set) are generated as an independent task, which is inserted to the beginning of the learning sequence. OWM with rejection category (OWM w/ rej.) significantly alleviates OODF, exhibiting in the relative accuracy drop $14.27\pm1.13$ is much lower than the OWM without rejection $44.11\pm2.42$ (OWM w/o rej.), in Tab.\ref{rebuttal_tab3}. While our primary focus in this work is to highlight the importance of OODF, we acknowledge that solving this problem is crucial. The approach presented here is a preliminary attempt, which will be thoroughly investigated in future.
	
	\section{Conclusion}
	
	In this work, we identify a new phenomenon of catastrophic forgetting, named out-of-distribution forgetting, and demonstrate how it can significantly affect the robustness of CL. Although OODF is described here in the image classification task under the class incremental scenario, it is straightforward to extend to other CL tasks in computer vision or natural language processing. OODF reveals the vulnerability of current CL methods in dealing with intra-class distribution shifts, which could be introduced intentionally or by unnoticed perturbations. This is well-conceivable in both attacking or accidental scenarios.
	
	More generally, our work suggests that the catastrophic forgetting problem in CL is more complex than we previously recognized, and it is likely that other forms of CF cannot be dealt with by the majority of current CL approaches may exist. Thus, it is of theoretical and practical importance to investigate the issue of CF more comprehensively, which will guide the development of more robust CL approaches that can work in complex environments.

	%
	
	\bibliographystyle{splncs04}
	\bibliography{Continual_Weakness4}
	
	%
	%
	%
	%
\end{document}